%% file: main.tex
\documentclass[12pt]{article}
\usepackage{amsfonts}
\usepackage{amsmath}
\usepackage{amsthm}
\usepackage[english]{babel}
\usepackage{booktabs}
\usepackage[labelfont=bf]{caption}
\usepackage{epigraph}
\usepackage{fullpage}
\usepackage{graphicx}
\usepackage[utf8]{inputenc}
\usepackage[numbers]{natbib}
\usepackage{parskip}
\usepackage{subcaption}

\usepackage[
  colorlinks=true,
  linkcolor=black,
  anchorcolor=black,
  citecolor=black,
  filecolor=black,
  menucolor=black,
  runcolor=black,
  urlcolor=black]{hyperref}

\DeclareMathOperator*{\argmin}{arg\,min}

\title{The Role of Evolution in Machine Intelligence}
\author{Awni Hannun\footnote{
  Send correspondence to
  \href{mailto:awni.hannun@gmail.com}{awni.hannun@gmail.com}}}

\date{\today}

\begin{document}
\maketitle

\defcitealias{pugh1977}{Emerson Pugh}
\setlength\epigraphwidth{.65\linewidth}
\epigraph{
\textit{If the human brain were so simple that we could understand it, we would
be so simple that we couldn't. }
}{\citetalias{pugh1977}}


\input{introduction}
\input{investing_in_phylogeny}
\input{phylogenetic_stack}
\input{starting_points}
\input{challenges}
\input{redistribution}

\section*{\large Acknowledgements}
I am grateful to Yusuf Hannun for feedback and comments which improved this work.

\bibliographystyle{plainnat}
\bibliography{references}

\end{document}

%% file: introduction.tex
\section{Introduction}

Two categories of development influence the behavior of an intelligent being:
(1) lifetime learning or \emph{ontogenesis} and (2) evolution or
\emph{phylogenesis}. The ontogenetic development of an organism includes
learning from the knowledge and experience accrued over a single lifespan. The
phylogenetic development of a species includes the knowledge accrued over the
course of evolution, which is passed from generation to generation through
the genome.

Research in developing intelligent machines can be categorized as either
ontogenetic or phylogenetic, as in table~\ref{tab:dev_categories}. The bulk of
current research efforts (2021) in the development of machine intelligence
belongs to the ontogenetic category. The primary phylogenetic development is
implicit in the lineage of research papers and software. In these artifacts we
see certain neural network architectures, optimization procedures, and
regularization techniques persist and evolve over time.

Evidence supports the importance of phylogenesis in the development of machine
intelligence. This evidence further suggests an underinvestment in research in
digital phylogeny. In order to understand our investment options, I describe
the ``phylogenetic stack''. This stack begins with the near \emph{tabula rasa}
environment of emergent evolution and proceeds to more assumption-filled
starting points culminating in purely ontogenetic machine learning. I assess
these various starting points by their ability to yield intelligent machines.
Since nature is the only system that has developed human-like intelligence, I
am limited to inductions from a sample size of one. Nevertheless, I rely on
natural evolution to guide in my analysis.

Ultimately, I aim to shed light on where in the phylogenetic stack to invest.
Given a limited research budget, we want to know which stage would yield the
largest return on investment, where the product is intelligent machines. My 
first-order suggestion is to diversify our research portfolio. We should invest
more broadly across the phylogenetic stack. My second-order hypothesis is
that meta-evolutionary algorithms (\emph{c.f.} section~\ref{sec:mea}) may yield
an optimal trade-off between the many factors influencing the development of
machine intelligence.

\begin{table}[t]
  \caption{A broad categorization of current approaches to develop machine
  intelligence as either ontogenetic or phylogenetic.}
  \centering
  \begin{tabular}{l | l}
  \toprule
  Ontogenetic Development & Phylogenetic Development \\
  \midrule
  Deep learning          & Architecture research \\
  Supervised learning    & Evolutionary algorithms \\
  Unsupervised learning  & Evolution of evolvability \\
  Reinforcement learning & Emergent evolution \\
  \bottomrule
  \end{tabular}
  \label{tab:dev_categories}
\end{table}

%% file: investing_in_phylogeny.tex
\section{Investing in Phylogeny}

Three motivations, not meant to be comprehensive, for increasing
investment in the phylogeny of machine intelligence are:

\begin{enumerate}
  \item The computation dedicated to natural evolution far exceeds that used in
	ontogenesis.
  \item Automation leads to less biased and more rapid progress.
  \item Intelligence is too complex to design by hand.
\end{enumerate}

\paragraph{Computation} The bulk of the computation which resulted in the
intelligence of humans occurred during evolution. The ontogenesis of an
individual leads to remarkable intelligence in only a few years, far surpassing
the best machines in generality. However, the genome, which allows for this, is
the result of nearly four billion years of evolutionary computation. Of course,
the types of computation are quite different. One may argue that the learning
during the lifetime of an individual is ``smart'' whereas the computation
occurring in an evolutionary process is ``dumb'', and hence the two cannot be
compared. However, the difference in time scales is approximately a factor of
$10^9$. Even considering the difference in efficiency between the two
computational processes, phylogenesis likely merits more computational
resources than are currently allocated to it.

\paragraph{Automation} Just as deep learning eradicated the need for
human-in-the-loop feature engineering, evolutionary algorithms should eradicate
the need for neural network architecture engineering. Similarly, traveling
further down the stack of computational models of artificial life will reduce
the need for hand-engineered evolutionary algorithms. Human-in-the-loop
algorithm improvement ({\emph i.e.} research) can also lead to bias which can
yield suboptimal results or hard to overcome local optima. Trend following
biases research towards exploitation of existing models rather than exploration
of new models and algorithmic niches. To further the analogy, the implicit
fitness function optimized by researchers may be littered with sub-optimal
critical points making the higher peaks of intelligence difficult to reach.

\paragraph{Complexity} We understand fairly well the low-level mechanisms which
participate in intelligent behavior. For example the cell is well understood
at the molecular level. We also understand the core units of the brain
including neurons, their connections, and how they transmit signals at a
granular level. However, we understand less about the decoding of the genome
and the complex network of interacting proteins which yield the central nervous
system capable of developing intelligence. We also understand little about the
high-level neural mechanisms from which intelligent behavior emerges.

We may be more successful developing machine intelligence as an emergent
property of a simpler-to-specify system rather than through more direct
human-in-the-loop design. In modern deep learning, we have surpassed the
threshold of explainable complexity. Researchers typically don't know why one
neural network architecture is more accurate than another. Furthermore, and
uncontroversially, the scale and complexity of current learning algorithms will
need to grow, possibly multiple orders of magnitude, before they exhibit a more
general intelligence.

\begin{figure}
  \centering
  \includegraphics[width=0.7\linewidth]{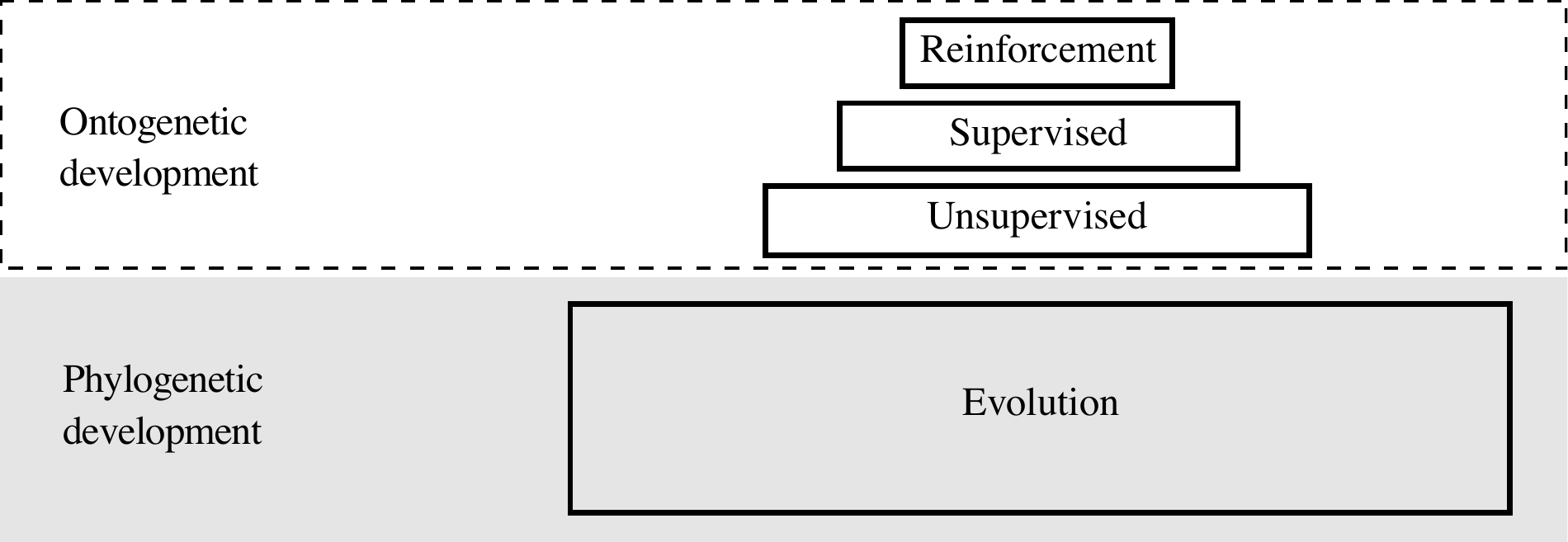}
  \caption{The layer cake of learning~\citep{lecun2016predictive} augmented to
  include the development due to evolution (phylogenesis). Current research is
  inverted with ontogeny receiving much more mind-share than phylogeny despite
  the later accounting for much more of the computation which lead to the
  emergence of intelligence.}
  \label{fig:layer_cake}
\end{figure}

The categories of machine-learning algorithms are sometimes
depicted as a layer cake~\citep{lecun2016predictive}.
Figure~\ref{fig:layer_cake} is one take on this layer cake analogy but with the
addition of evolution and the ontogenetic and phylogenetic stages of
development demarcated. The size of each layer represents the relative amount
of data or computation needed to achieve intelligent behavior.  Research and
resource allocations can draw inspiration from figure~\ref{fig:layer_cake}. The
implication is that we are underinvested in the phylogenetic development of
machine intelligence.

%% file: phylogenetic_stack.tex
\section{The Phylogenetic Stack}
\label{sec:phylo_stack}

\begin{figure}
\centering
\includegraphics[width=0.73\textwidth]{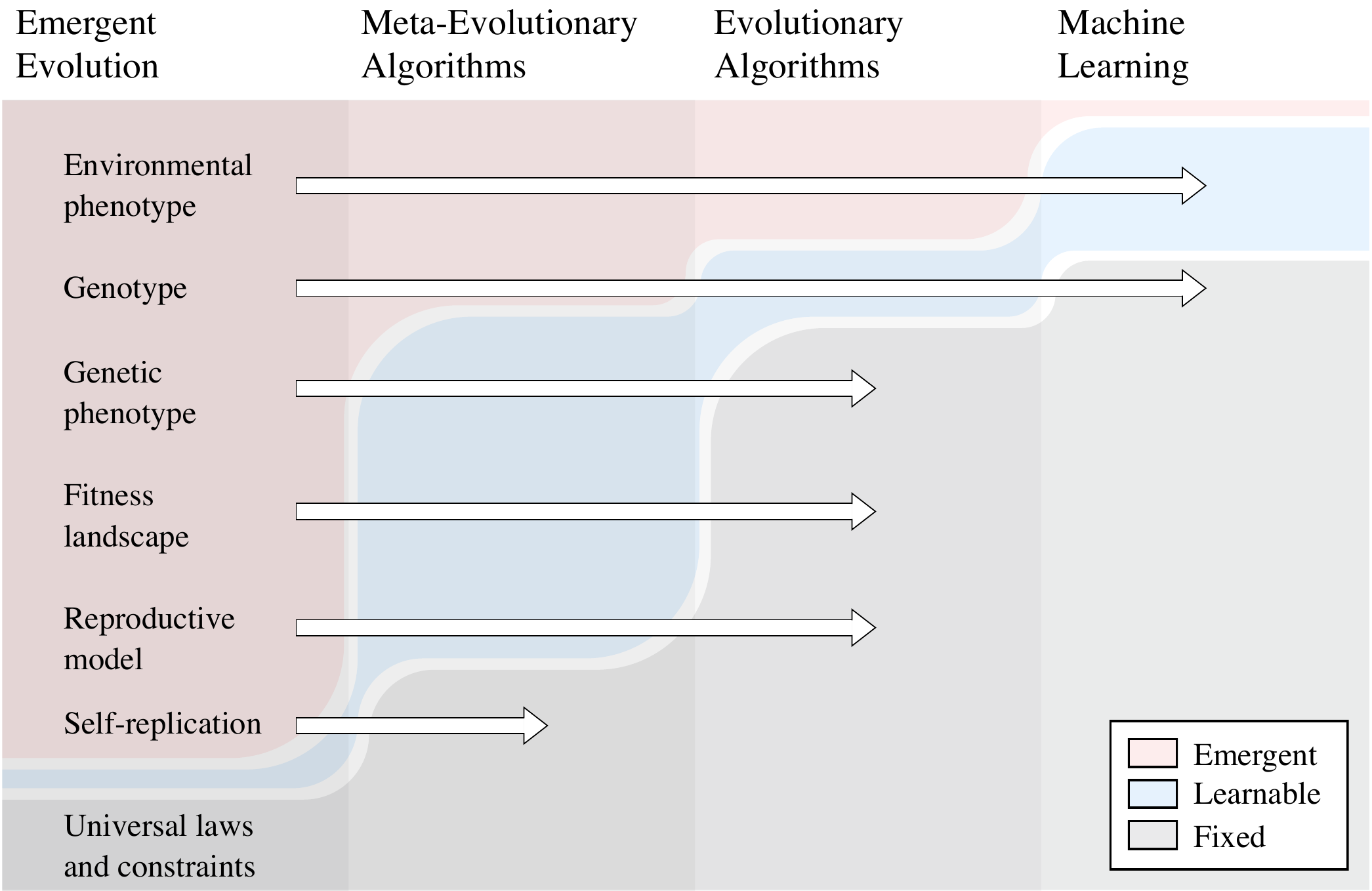}
\caption{Four starting points in the phylogenetic stack are shown horizontally
    at the top in order of increasing assumptions required. Stacked vertically
    on the left are many of the core modeling components present (explicitly or
    implicitly) in every evolutionary process. The colored regions indicate
    whether the model is emergent, learnable, or fixed. The arrows indicate how
    a given model changes regions across the phylogenetic stack.}
\label{fig:model_cascade}
\end{figure}

We describe four potential starting points from which to develop intelligent
machines beginning near the bottom of the phylogenetic stack and culminating with
purely ontogenetic development. These starting points are: 1) emergent
evolution, 2) meta-evolutionary algorithms, 3) evolutionary algorithms, and 4)
machine learning. For each starting point, we discuss common models of
computation and some of the key assumptions required for starting at that
level. While we have quantized the phylogenetic stack into four starting
points, the space between these points is not empty. Other, potentially more
optimal starting points may exist between them. Figure~\ref{fig:model_cascade}
shows a summary of the four starting points, and their relationships to the
core modeling components explicit or implicit in any evolutionary process.

\subsection{Emergent Evolution}
\label{sec:emergent_evo}

In emergent evolution, the goal is to setup a dynamical system in which an
evolutionary process emerges. In particular, the designer may not specify the
rules of replication or, for example, explicitly seed the environment with
self-replicating structures. Instead, the designer specifies a simpler
computation model from which an evolutionary process can emerge.
Following~\citet{dennett2002new}, we define an evolutionary process as
consisting of individuals or groups of individuals which exhibit
self-replication with potential mutations in a competitive environment.

Some example computation models include artificial
chemistries~\citep{dittrich2001artificial} and cellular automata
(CA)~\citep{neumann1966theory}. In both approaches a state space and a
transition model must be specified. The state space includes a description of
the current set of individuals as well as any additional resources. The
transition model specifies the rules which govern the propagation of the state
from one time to the next. These two broad components include many important,
nearly axiomatic, properties of the system. For example, resource constraints
are included in the state space and conservation laws may be implicitly
included or explicitly specified in the transition model.

One of the challenges with large-scale assumption-free simulations is simply
recognizing the emergence of interesting behavior. The existence of
self-replicating individuals and other features of an evolutionary process may
not be immediately apparent. In order to recognize interesting behavior in
these systems one must define carefully what it means to be ``interesting'' and
create tools capable of measuring this. For example,
compression-based~\citep{zenil2010} and prediction-based
metrics~\citep{cisneros2019evolving} can be used to measure the complexity of
the states generated by a cellular automata. One challenge is finding measures
which capture the zone between the extremes of trivial and purely random
behavior, both of which are undesirable.

Emergent evolution obviates the need to specify by hand difficult-to-design
processes which are likely to be critical to the development of machine
intelligence. For example, the ``evolution of
evolvability''~\citep{dawkins1988} may naturally emerge from a model which
yields an evolutionary process. This improvement in the rate of evolution may
be important to the emergence of intelligent behavior. Similarly, while
indirect encodings can be used in traditional evolutionary
algorithms~\citep{cangelosi1994cell, stanley2007compositional,
stanley2009hypercube}, and genotype-to-phenotype mappings may even be
learned~\citep{moreno2018learning}, the space of such mappings can be explored
with less bias by starting further down in the phylogenetic stack.

\subsection{Meta-Evolutionary Algorithms}
\label{sec:mea}

We use the term ``meta-evolutionary algorithm'' (MEA) to refer to an
evolutionary algorithm (EA) for which the usually fixed components can
themselves be evolved. Any EA must specify at least a genotype-to-phenotype
mapping (or use a direct encoding), an evaluation process (typically a
hand-designed fitness function), and a reproduction process (typically mutation
combined with crossover).  Usually these models are fixed and only the genotype
is adapted. A partial MEA allows at least one of these components to evolve
along with the rest of the algorithm. A complete MEA allows all of them to
evolve.

The distinction between MEAs and emergent evolution (\emph{c.f.}
section~\ref{sec:emergent_evo}) is that an MEA may seed the initial models.
The initial population may already possess the ability to self-replicate as
well as a preliminary genotype-to-phenotype function. In emergent evolution
these abilities must emerge from the simulation.

A benefit of MEAs over emergent evolution is that they retain the ability to
modify the key components of phylogenesis (replication, recombination,
\emph{etc.}) without the computational burden required for their discovery.
Arguably the emergence of these components is not the critical aspect of the
ability of phylogenesis to yield intelligence. Perhaps more important is simply
their adaptability.

The umbrella category of MEAs encompasses concepts such as the ``evolution of
evolvability'' (evo-evo)~\citep{dawkins1988}, evolutionary development
(evo-devo) when applied to learning genotype-to-phenotype mappings, and other
intrinsic (as opposed to extrinsic or hard-coded)
processes~\citep{packard1989intrinsic, taylor2019evolutionary}. These have all
been active topics in the evolutionary biology literature for decades but are
not yet central to the application of EAs in developing machine intelligence.

As in emergent evolution, some degree of open-endedness may be important to
allow the system to develop sufficiently interesting behavior. Similarly,
measuring the complexity of the evolved organisms will likely be necessary to
identify non-standard yet intelligent behavior. We elaborate on these common
challenges in section~\ref{sec:challenges}.

\subsection{Evolutionary Algorithms}

Evolutionary algorithms (EAs) have thus far led to limited improvements in
state-of-the-art machine learning. They have been used over the past several
decades to tune model architecture and hyper-parameters for a given
machine-learning problem~\citep{miikkulainen2019, miller1989, real2019,
stanley2002evolving}. However, these results are mixed, with little adoption in
practice. Manual tuning, grid search, random search, and Bayesian optimization
make up the bulk of currently used methods~\citep{bouthillier2020survey}.
Genetic algorithms, a subset of EAs, are widely used. They are especially
useful when solving optimization problems over large discrete spaces, which are
not amenable to first and higher-order gradient-based
methods~\citep{mitchell1998}.

Three key components of an EA are 1) the genotype encoding and corresponding
mapping to the phenotype (or lack thereof for direct encodings), 2) the fitness
function, and 3) the reproduction process which includes crossover and mutation.
Following the discussion in section~\ref{sec:mea}, we refer to these as the
meta-models of the EA.

The sub-optimality of the meta-models could account for the underutilization of
EAs in machine-learning. Human-in-the-loop design usually performs better.
Continuing to improve these models directly is one route to enable the
development of machine intelligence with EAs. Early research in indirect
genotypes for evolving modular neural networks showed promise in learning
complex functions~\citep{gruau1993genetic}. In comparison to a direct encoding,
the indirect encoding solved more difficult problems much
faster~\citep{gruau1996comparison}. This and related approaches are worth
revisiting at the much larger scale attainable with today's hardware and data.

Alternatively, taking a step down the phylogenetic stack to MEAs could be more
effective. Whether MEAs or human-in-the-loop design is the best path
for finding more optimal genotype-to-phenotype mapping, fitness functions, and
reproduction models depends on the computational burden of discovery. Hybrid
approaches are also possible.

Regardless of the approach, more sophisticated meta-models will likely be
critical to the phylogenesis of machine intelligence. For example, a directly
encoded genotype will scale poorly to the complexity of the underlying
phenotype. All organisms in nature indirectly encode their phenotype in DNA.
The mapping from the genotype to the phenotype involves a sophisticated
regulatory network dictating gene expression. This process plays a key role in
enabling the complex phenotypes of plants and animals, let alone human
intelligence.

Another possibility is that EAs are missing one or more key ingredients other
than the three described above. These components might be necessary for a
continual push towards ever greater complexity or open-endedness. We discuss
the challenge of complexity and open-endedness, which is common to most of the
phylogenetic stack, in section~\ref{sec:open_endedness}.

\subsection{Machine Learning}
\label{sec:machine_learning}

The three pillars of machine learning are supervised learning, unsupervised
learning, and reinforcement learning (\emph{c.f.} figure~\ref{fig:layer_cake}).
These three approaches are often used in conjunction with one another; representing
a continuous space of techniques rather than three isolated points. For example,
semi-supervised learning techniques lies on the line between fully supervised and
unsupervised learning. Similarly, reinforcement learning often relies on
imitation learning in which a high value action from a given state is provided
by an expert~\citep{schaal1999imitation}. This amounts to learning from labeled
data, yielding a hybrid between fully supervised and reinforcement learning.

A natural question to ask is if exploring the simplex spanned by these three
techniques is likely to lead to the development of human-like machine
intelligence. If not, we should attempt to understand the missing ingredients.

Focusing exclusively on these three pillars of machine learning is too narrow.
Machine-learning researchers study a broad array of related problems and
approaches. These include many subfields such as life-long (or continual)
learning, multi-task learning, active learning, few-shot learning, domain
adaptation, transfer learning, and meta-learning. These subfields typically aim
to enable an approach based on one of the fundamental learning pillars to
generalize to more tasks (life-long or continual learning, and multi-task
learning) or to more rapidly generalize to a given task (active learning and
k-shot learning), perhaps by taking advantage of previous knowledge (domain
adaptation, transfer learning, and meta-learning).

Improving the rate of generalization of machines to a given task and enabling
generalization to more tasks are arguably the most important missing criteria
for machines to exhibit human-like intelligence. However, Tesler's
theorem~\citep{tesler} and his corresponding dictum ``intelligence is whatever
machines have not done yet'' (often referred to as the AI
Effect~\citep{haenlein2019brief}) hint that developing human-like machine
intelligence may be more subtle.

One possibility for the missing link is enabling some degree of open-ended
behavior by the machine~\citep{mikolov2016roadmap}. Generative
adversarial models~\citep{goodfellow2014} and reinforcement learning models
which engage in self-play~\citep{silver2018, tesauro1995temporal} are a step
towards open-endedness~\citep{guttenberg2019potential}. Assuming fixed data
distributions and model hypothesis spaces, these models usually reach a fixed
point. Drift in the complexity of the data distribution and hypothesis space
may enable a larger degree of open-endedness.

%% file: starting_points.tex
\section{Assessing Starting Points}
\label{sec:starting_points}

Research in the phylogenetic stage of development can be thought of as moving
further down the stack in the computational infrastructure giving rise to
intelligence. The further down the stack we go, the fewer assumptions we need
to make. This is beneficial. The more assumptions about the underlying
computational model and algorithms we make, the more likely they are to be
incorrect and hence limiting to the emergence of intelligence. However, the
further down the stack we go, the more we rely on undirected but interesting
behavior to emerge. This likely requires substantially more computation since
much of the computation will be wasted on uninteresting behavior. In this case,
we also require some notion of ``interesting'' and an ability to identify
behavior as such. In a massive simulation, merely determining if an
evolutionary process has emerged can be difficult. Furthermore, identifying the
intelligent behavior itself can also be problematic. In an undirected
assumption-free simulation, emergent intelligence may be incomparable to that
which we are used to observing in animals and humans. I describe in more
detail the challenges and paths forward for identifying and measuring
intelligence and the more nebulous ``complexity'' in
section~\ref{sec:identifying_intelligence}.

\begin{figure}
\centering
\includegraphics[width=\textwidth]{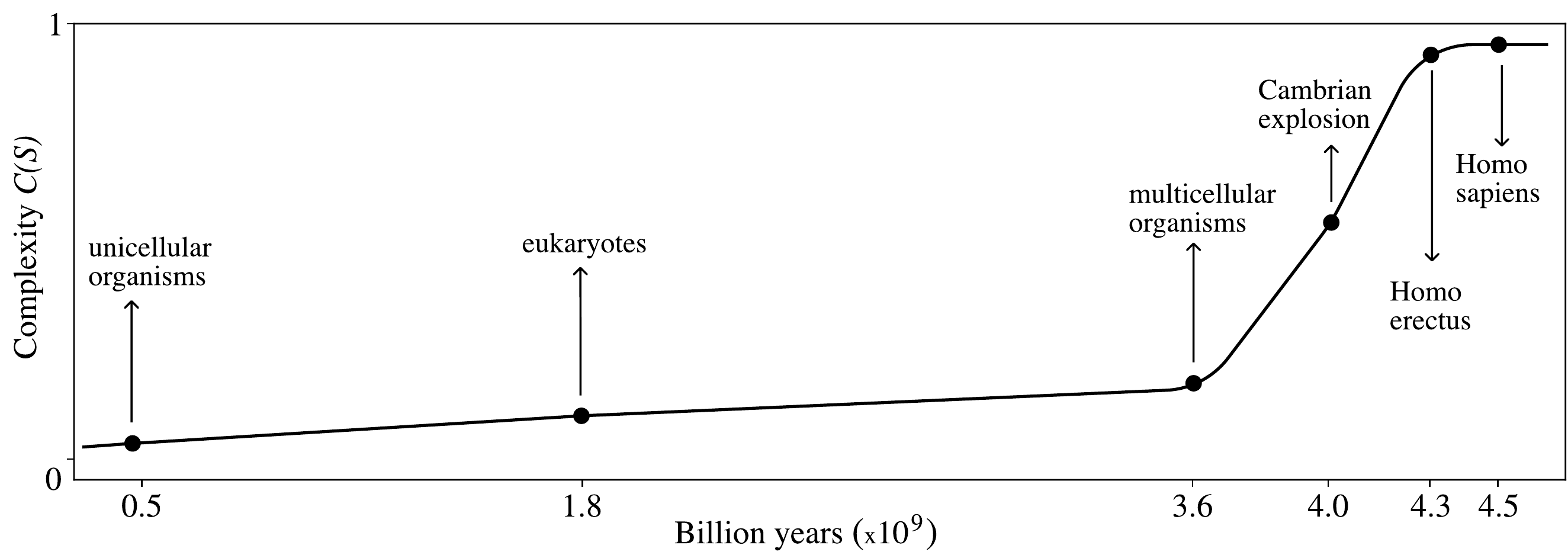}
\caption{On the horizontal axis is a timeline of the evolution of intelligent
  life in billions of years since the formation of Earth. Certain landmarks are
  indicated, such as the emergence of multicellularity or the Cambrian
  explosion. For each point on the timeline, we provide an estimate of the
  complexity of the most advanced organisms.}
\label{fig:understand_vs_life}
\end{figure}

Deciding where in the phylogenetic stack to invest in research requires
navigating complex trade-offs. I explore these trade-offs from a few angles.

\subsection{Evolutionary Time and Complexity}
\label{sec:time_vs_complexity}

The complexity of the organisms present at a given stage of evolution can help
determine where to start in a digital simulation. As the complexity of the
organism grows, determining which properties are important to encode \emph{in
silico} and how to encode them becomes more challenging. This suggests starting
with as simple a system as possible. However, the simpler the starting point,
the greater the computational time required to yield human-like intelligence.

We observe the relationship between complexity of various organisms and the
time required to evolve human intelligence from the first appearance of that
organism. This can guide our choice of where to start in the phylogenetic
stack. Assume the quality of a starting point $S$ is proportional to the sum of
the complexity of the most advanced organism at that point $C(S)$ and the
time-to-intelligence $T(S)$ (where both $C(S)$ and $T(S)$ are appropriately
scaled). We look for the starting point by finding the minimizer:
\begin{equation}
  S^* = \argmin_S C(S) + T(S).
\end{equation}
Upon finding $S^*$, we can evaluate the evolutionary models present at that
point. This evaluation can inspire where in the digital phylogenetic stack
(\emph{c.f.} figure~\ref{sec:phylo_stack}) to begin based on which models are
emergent, learned, or fixed.

Figure~\ref{fig:understand_vs_life} shows a timeline of the evolution of
intelligence since the formation of Earth. At each point we estimate the
complexity of the most advanced organism present at that time. The complexity
in figure~\ref{fig:understand_vs_life} is based on the judgement of the author
and is not intended to be definitive. A more quantitative approach could
use for example the size of the genome. However, this suffers from the
well-known C-value paradox~\citep{thomas1971}.~\footnote{A similar problem arises,
aptly dubbed the G-value paradox, when using the number of genes instead
of the C-value (mass of DNA)~\citep{hahn2002g}.} Another approach could be
to estimate the information carrying capacity in bits of the central nervous
system of the most advanced organism. \citet[p. 26]{sagan2012dragons} gives an
example of both approaches, yielding curves with a similar shape as that
of figure~\ref{fig:understand_vs_life}.

Interestingly, figure~\ref{fig:understand_vs_life} has a knee shape with a bend
roughly around 3.5 billion years at the development of macroscopic
multicellular organisms.\footnote{This is an estimate for the emergence of
macroscopic multicellular organisms, however; evidence exists that smaller and
simpler multicellular organisms emerged for the first time more than 2 billion
years ago~\citep{han1992megascopic}.} The development of macroscopic
multicellular organisms was soon followed by the Cambrian explosion and the
genus Homo. However, while unicellular organisms developed soon after the
cooling of the Earth, multicellular organisms are believed to have taken much
longer. This point in evolutionary history may be difficult to discover by a
more assumption-free evolutionary process. On the other hand, we understand a
large portion of the processes which govern simple multicellular organisms.
They are considerably less complex than the many larger organisms that evolved
during and after the Cambrian explosion. Because of this, simulation of an
evolutionary process starting at a digital analog of multicellularity may yield
an optimal trade-off in the amount of direct design required and computation
needed to observe emergent intelligence.

The next step is to determine the digital analog of evolution at the time of
the initial macroscopic multicellular organisms. When multicellular organisms
emerged, nature was already equipped with several key components of an
evolutionary process. The environment itself consisted of finite resources
engendering competition and natural selection under complex fitness landscapes.
Also present was a large diversity of self-replicating organisms such as the
eubacteria, archaea, and both of the unicellular eukaryotes (protozoans) and
multicellular eukaryotes (metazoans). All of these organisms had (and have) a
common method to store information (DNA), and a related genotype-to-phenotype
decoder (RNA, ribosomes, proteins, \emph{etc.}).  Mutation and recombination
existed at the very least via random errors produced during DNA cloning and
through exchanging DNA in plasmids. However, many of these core components
continued to develop including crossover and sexual reproduction, more
sophisticated machinery for decoding genes, and more complex fitness landscapes
due in part to feedback from the growth of biotic life in the environment.

Overall, these observations point to a meta-evolutionary algorithm as a good
starting point. An MEA can yield an optimal trade-off between wasted
computation spent on the emergence of evolutionary behavior and artificial
barriers to machine intelligence from poorly specified models.

\subsection{Computation and Assumptions}
\label{sec:compute_vs_assumptions}

The number of assumptions required at each starting point can serve to guide
the decision of where to begin the development of a computational model leading
to intelligence. Fewer assumptions imply a larger state space of results
achievable by the simulation. This decreases the chance that the designer
inadvertently removes areas in the state space where intelligent behavior is
likely to develop. The trade-off is that exploring the larger space requires
more computation.

We define an assumption as any modeling choice which narrows the state space
that the simulation would have had access to without the assumption. From this
definition we can also assign a magnitude to each assumption based on the
amount that it shrinks the search space of the simulation compared to that of
the assumption-free version. Let $A(S)$ be the cumulative magnitude of the
assumptions for a given starting point, $S$, and $T(S)$ be the computation time
required to develop machine intelligence. A good starting point is one which
minimizes the sum of the properly scaled assumptions and computation time:
\begin{equation}
  S^* = \argmin_S A(S) + T(S).
\end{equation}

\begin{figure}
\centering
\includegraphics[width=\textwidth]{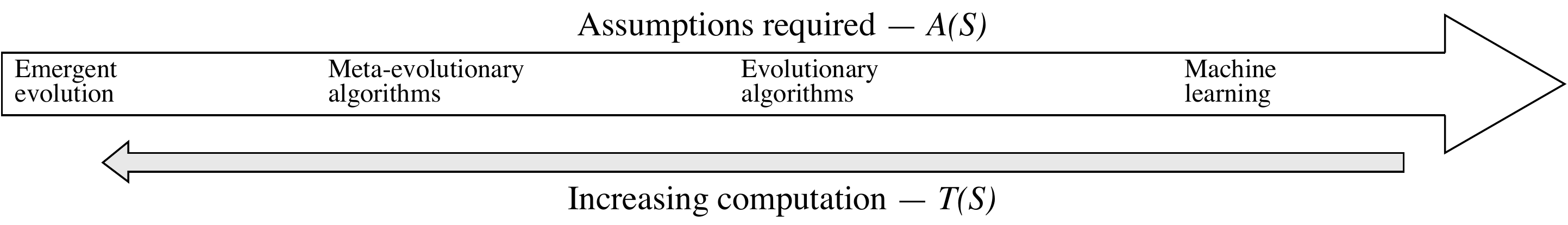}
\caption{Selected starting points of the development stack ordered by the
    number of assumptions needed. The fewer assumptions needed the more
    computation.}
\label{fig:assumptions}
\end{figure}
Figure~\ref{fig:assumptions} depicts four of the possible starting points in
the phylogenetic stack ordered by increasing magnitude of assumptions required.
The bottom arrow indicates that the computation time, $T(S)$, increases in the
opposite direction.

Emergent evolution requires the fewest assumptions. Possibly any model
computationally equivalent to a universal Turing machine is sufficient for
intelligent behavior to eventually emerge. Wolfram's ``Principle of
Computational Equivalence''~\citep{wolfram2002new} suggests that such Turing
complete models are abundant in nature. Even the simplest digital systems such
as elementary cellular automata can be computationally
universal~\citep{cook2004universality}. While these universal systems are all
equivalent in what they can simulate, they may not be equal in the likelihood
and amount of computation required to evolve intelligent behavior. Hence, the
assumption of which class of models to use and selection of the specific model
within that class may be critical.

Meta-evolutionary algorithms and evolutionary algorithms require strictly more
assumptions than a model of emergent evolution. In MEAs, the self-replicating
individuals and a core set of meta-models (a genotype-to-phenotype map, a
reproductive model, and a fitness landscape) are pre-specified but adaptable.
The assumptions implicit in EAs are greater in magnitude than MEAs as the
meta-models themselves do not adapt -- only the genotype is evolved.

Machine learning requires a greater cumulative magnitude of assumptions than an
evolutionary algorithm. As an example, in a typical use of deep neural
networks, the designer specifies the model architecture, learning criterion,
optimization algorithm, and regularization. An EA operates on a family of model
architectures, whereas a machine-learning algorithm operates on a single,
typically hand-designed architecture. If the EA uses an expressive indirect
encoding of the phenotype (architecture) then the relative magnitude in
assuming a single architecture is even greater.

Identifying interesting behavior becomes more difficult with fewer assumptions.
In a simulation of artificial life the goal may be to observe emergent
evolution and intelligent behavior. Identifying self-replicating individuals
undergoing mutation and selection may be difficult in a massive simulation.
Intelligent behavior may take an unfamiliar form and can be even more difficult
to identify. This requires care in specifying and measuring
intelligence and any other types of complex behaviors we intend to observe.

\subsection{Contingency and Convergence}

The prior sections asses starting points based on the initial complexity and
magnitude of the assumptions required. Optimal starting points are those
which minimize the trade-off between these and the computation time required to
evolve intelligence. We used natural evolution as a guide for the computation
time in section~\ref{sec:time_vs_complexity} and left computation time mostly
unspecified in section~\ref{sec:compute_vs_assumptions}. This estimate of
computation time does not address the potential contingency of human-like
intelligence in natural evolution. Perhaps nature was simply lucky and evolved
intelligence much more quickly than expected. Or perhaps nature was unlucky and
intelligence could have evolved much more quickly. We attempt to understand on
average how much time would be required to evolve human-like intelligence. This
can then guide where in the phylogenetic stack to initiate a digital
evolutionary system.

Throughout this section we rely on nature to support our arguments. An
important consideration is which of the processes found in nature are necessary
to translate into a digital simulation and which parts are artefacts of the
natural world and irrelevant to the digital analog. We discuss this
consideration in more detail in section~\ref{sec:balance_bio}.

A contingent phenotype (or species, phylum, \emph{etc.}) in evolution is one
which, while possibly predictable or explainable in hindsight, is not probable.
The contingent phenotype is one of many likely alternatives that could have
occurred subject to very slight deviations in initial conditions or external
influences. \citet{gould1990wonderful} suggests that contingency plays a large
role in evolution, citing as evidence the Cambrian radiation of diverse phyla
followed by a seemingly arbitrary culling. By this logic the development of
humans may be highly unlikely to occur again if we were to replay the course of
evolution. However, this does not imply that intelligence itself is unlikely.

An evolutionary niche dictates the likelihood of a phenotype emerging based on
two factors: 1) the strength of selection for the given phenotype in the niche
and 2) the capacity of the niche. The strength of selection for a phenotype is
dictated by how much it benefits an organisms ability reproduce. A higher
strength of selection within a niche leads to \emph{convergent} evolution.  The
evolution of a highly convergent phenotype in a given niche will be robust to
initial conditions and external perturbations. The capacity of the niche
dictates the degree of \emph{parallel} evolution. The same phenotype can evolve
in different organisms if the niche has a high capacity for it.  Together these
characteristics of an evolutionary niche dictate the level of contingency or
likelihood of observing a given phenotype.

We observe with some confidence that nature does not have a high capacity for
human-like intelligence. It has evolved only once in history and does not appear
to be evolving in parallel in any contemporary species. Estimating the degree
of convergence of intelligence is more difficult. \citet{morris2003life} argues
that evolution is highly convergent and human-like intelligence is not only
predictable but probable.\footnote{Somewhat ironically, Conway Morris is one of
the three primary re-discoverers of the Burgess shale which is the pillar that
\citet{gould1990wonderful} builds his case of contingency around.}

As examples, we consider the contingency of two possible ``major transitions''
in natural evolution~\citep{smith1997major}. First consider the hypothesized
endosymbiotic merger of the archaea and eubacteria which resulted in the modern
eukaryote with its energy generating mitochondria~\citep{martin1998hydrogen}.
\citet{lane2006power} argues that this merger is extremely contingent and
unlikely to be a convergent result in evolution (and therefore also
unlikely to evolve in parallel). \citet{lane2006power} also argues that the
merger of the two prokaryotes was a necessary condition for the growth of the
genome\footnote{The argument goes roughly as follows. In order to grow beyond a
certain limit, cellular energy production must scale linearly with cell volume.
Bacterial energy production uses the cell membrane to create the proton
gradient required for respiration. This scales with surface area and not
volume, and hence puts a limit on cell size. Local organelles which produce
energy throughout the volume of the cell enable linear scaling of energy
production with the volume of the cell. However, in order to regulate energy
production at a granular level, the mitochondria need local genetic information
(the co-location for redox regulation
hypothesis~\citep{allen2015chloroplasts}). The simplest (and perhaps only
plausible) way for this to evolve is if one cell engulfs another.} and hence
the development of more complex multicellular organisms and ultimately
human-like intelligence.

Another major transition is the evolution of multicellular organisms from
unicellular organisms. Multicellularity has evolved independently at least 25
times~\citep{grosberg2007evolution}. The spontaneous self-organization of cells
into larger groups appears to be highly convergent in natural evolution.
This suggests that multicellularity has a strong reproductive benefit and that
discovering it in an evolutionary process is not so difficult. As far as
digital evolutionary systems go, encoding modularity as an analog to
multicellular organisms may benefit evolvability. However, even if one does not
directly encode such a phenotype, the convergence in natural evolution suggests
that this may also emerge naturally in a sophisticated simulation.

%% file: challenges.tex
\section{Challenges in Evolving Intelligence}
\label{sec:challenges}

\subsection{Biological Relevance}
\label{sec:balance_bio}

Thus far we have somewhat egregiously conflated natural evolution with digital
evolution. Nature has certainly inspired the design of digital
algorithms, a few of which include genetic algorithms, neural networks, and swarm
intelligence~\citep{fister2013brief}. However, carelessly following
nature's path may be suboptimal or even misleading.

Consider flight which has evolved independently four times across four major
animal classes.~\footnote{Flight evolved independently in mammals (bats),
birds, insects and reptiles (pterosaurs).} Flying in nature is thus relatively
convergent. Despite the propensity for evolution to discover flight, directly
designing airplanes using the principles of aeronautics is a better approach.
Alternatively, consider the wheel. The wheel is simple to design but is thought
to have evolved in nature only once, namely in the flagellum used for forward
locomotion by prokaryotes~\citep{labarbera1983wheels}.

Human-like intelligence falls in a different regime than either of the above
examples. It is neither simple to design directly nor is it highly convergent
in evolution. Which route is easier, or even feasible, is yet to be determined.
Even if human-like intelligence is easier to discover with evolution, a direct
approach may yield a more efficient and usable design.

Many processes or phenotypes in natural evolution may be important to drive the
complexity and diversity needed to ultimately yield intelligent behavior.
Consider a few examples:

\begin{itemize}

  \item Non-coding DNA may serve as a substrate for mutation which is more
    likely to yield non-deleterious mutations. We may need analogous scratch
    space in the digital genotype.

  \item Multicellularity and other kinds of modularity could be important for
    the ``evolution of evolvability''~\citep{dawkins1988}. Incorporating this
    in any direct encoding of an evolutionary algorithm may yield more rapid
    and responsive evolution.

  \item Genetic crossover and sexual reproduction may be essential to enable
    exploration in evolution while avoiding error
    thresholds~\citep{biebricher2005error}. Similarly, in emergent evolution we
    may need to enable approaches which can avoid cascading errors and overcome
    Eigen's paradox\footnote{Eigen's paradox argues that given the mutation
    rates found in nature, without error correction during DNA self-replication
    the largest possible genome sizes are quite limited. This size limitation
    is much smaller than the size of a genome which can encode an error
    correction mechanism.} while maintaining a healthy mutation
    rate~\citep{eigen1971}.

\end{itemize}

The features of natural evolution that can improve digital evolution must
ultimately be determined empirically. However, qualitatively assessing the
important features of natural evolution and their ability to translate to
useful digital analogs can help to prioritize research in digital evolution.
After all, nature is the only system we know of that has created human-like
intelligence.

\subsection{Complexity and Open-endedness}
\label{sec:open_endedness}

A drive towards ever greater complexity is arguably one of the key missing
criteria from digital evolutionary and learning systems. In natural evolution,
the increase in peak complexity is hard to refute despite the term being poorly
defined and difficult to measure. A key challenge is understanding what drives
evolution's ``arrow of complexity'' and how to achieve a similar effect in
artificial systems~\citep{bedau2008arrow}.

A related notion is that of open-endedness, of which unbounded complexity
growth is a defining characteristic. Recently, open-endedness in artificial
systems has become a research goal in and of itself~\citep{stanley2017open}.
This has not been without controversy, as coming up with definitions of
open-endedness which do not admit trivial solutions has been a
challenge~\citep{hintze2019open, taylor2020importance}.

Both open-endedness and complexity are not well-defined terms with disagreement
as to their importance and how to measure them. Regardless, some degree of an
open-ended drive toward complexity is likely to be important to the emergence
of machine intelligence from any of the starting points discussed in
section~\ref{sec:starting_points}. We discuss some possible enablers and
drivers of complexity and open-endedness in the context of these starting
points.

\paragraph{Emergent evolution} One driver of complexity in emergent evolution
is the hierarchical construction of higher-level components from lower-level
building blocks. For example, in computational chemistries, monomers are
constructed from atoms, polymers from monomers, and more complex and
even self-replicating molecules from groups of polymers.
\citet{rasmussen2001ansatz} propose that increasing the complexity of the
primitives of such a system is a necessary and sufficient condition for the
emergence of higher-order structure. This is related to the generation of
open-ended evolution through ``cardinality
leaps''~\citep{sayama2019cardinality}. A cardinality leap can be achieved
through hierarchical construction. For example, a finite set of primitive
building blocks can be used to construct a countably infinite number of
higher-level structures.

The emergence of higher-level structure from lower-level primitives seems
likely to be a key driver of complexity growth in a simulation of emergent
evolution. However, complexity on its own is not a sufficient condition for the
emergence of self-reproducing organisms. Other properties are likely to be
important, including a tendency towards entropy, resource constraints, and
possibly other conservation laws.

\paragraph{EAs and MEAs} Many theories have been proposed to explain the
natural tendency of evolution to grow the complexity of organisms. Some of
these include:

\begin{itemize}
  \item The theory of ``passive diffusion'' posits that organisms in
    evolutionary systems grow in complexity purely due to randomness. In a
    random walk, the probability of arriving at a larger distance from the
    origin increases with the number of steps. In same way, the probability of
    developing more complex organisms increases over
    time~\citep{mcshea1994mechanisms, gould1996full}.

  \item Red queen effects~\citep{van1973new} due to predator-prey arms races and
    other inter-species coevolutionary dynamics are hypothesized to be
    drivers of complexity~\citep{dawkins1997climbing}.

  \item \citet{korb2011evolution} state that complexity of an evolutionary
    system (as opposed to organisms) can grow through niche construction. They
    argue that a given niche creates multiple byproducts which in turn support
    the development of more than one new niches resulting in exponential
    growth. We hypothesize in turn that an increase in evolutionary niches can
    yield more complex coevolutionary dynamics which could in turn yield a
    growth in organism complexity.
\end{itemize}

These are but a few plausible explanations for complexity growth in
evolutionary systems. Furthermore, these explanations do not preclude one
another. Incorporating one or more of these drivers of complexity growth in an
EA or MEA will likely aid in the emergence of intelligence.

Evolutionary systems also have features which suppress complexity. Thus
avoiding complexity suppressors in a system is also important. For example,
natural selection requires mutation, but if the rate of mutation is too high it
may induce an error threshold on the length of the
genome~\citep{biebricher2005error}. Diversity while avoiding error thresholds
can be achieved through other mechanisms such as crossover or gene duplication.
However, point-wise mutations may be important to explore the evolutionary
landscape at a finer grain.

\paragraph{Machine learning} The complexity of a machine-learning model is
dictated by the hypothesis space and the objective function. Following Occam's
razor, most learning algorithms prefer simpler over more complex models.
Regularization allows practitioners to trade-off fitness to the learning
objective for simplicity with the intention that this yields models which
generalize better to the true data distribution. The reasonable desire to have
models which generalize acts a suppressor of complexity.

Growing complexity in machine learning simply for sake of complexity
does not make sense. The driver of complexity must come from more diverse,
difficult to achieve, and possibly open-ended learning objectives.

Recently a trend has emerged in machine learning research to construct
benchmarks with multiple tasks~\citep{wang2018glue}. Models have been developed
which perform well on these benchmarks with minimal task-specific
training~\citep{devlin2018bert}. This includes models which are able to perform
tasks across differing modalities, such as vision and text~\citep{lu2020}. As
the size, number, difficulty, and diversity of the tasks in these benchmarks
grow, so too will the required complexity of a model which can generalize
across the full suite.

While encouraging complexity, the trend towards multi-task generalization does
not require open-ended learning. As mentioned in
section~\ref{sec:machine_learning}, open-ended machine learning systems may be
important to developing more human-like
intelligence~\citep{mikolov2016roadmap}. Steps towards achieving this type of
open-ended behavior include for example generative adversarial models and
self-play in reinforcement learning~\citep{guttenberg2019potential}.

\subsection{Identifying Intelligence}
\label{sec:identifying_intelligence}

The further down the phylogenetic stack we begin an evolutionary simulation the
fewer assumptions we need. These assumptions will have a tendency to lead to
more familiar types of intelligent behavior. Conversely, fewer assumptions will
make identifying emergent intelligence (or the emergent ability to rapidly
evolve intelligence) difficult.

Assuming a close correspondence of intelligence and complexity, one way to
identify a system capable of emergent intelligence is to measure complexity.
Researchers have attempted to qualitatively assess the features of systems
which exhibit complexity~\citep{wiesner2019measuring}  However, as
simulations grow in size and number, qualitative examination is not scalable.

\begin{figure}
  \centering
  \includegraphics[width=0.8\textwidth]{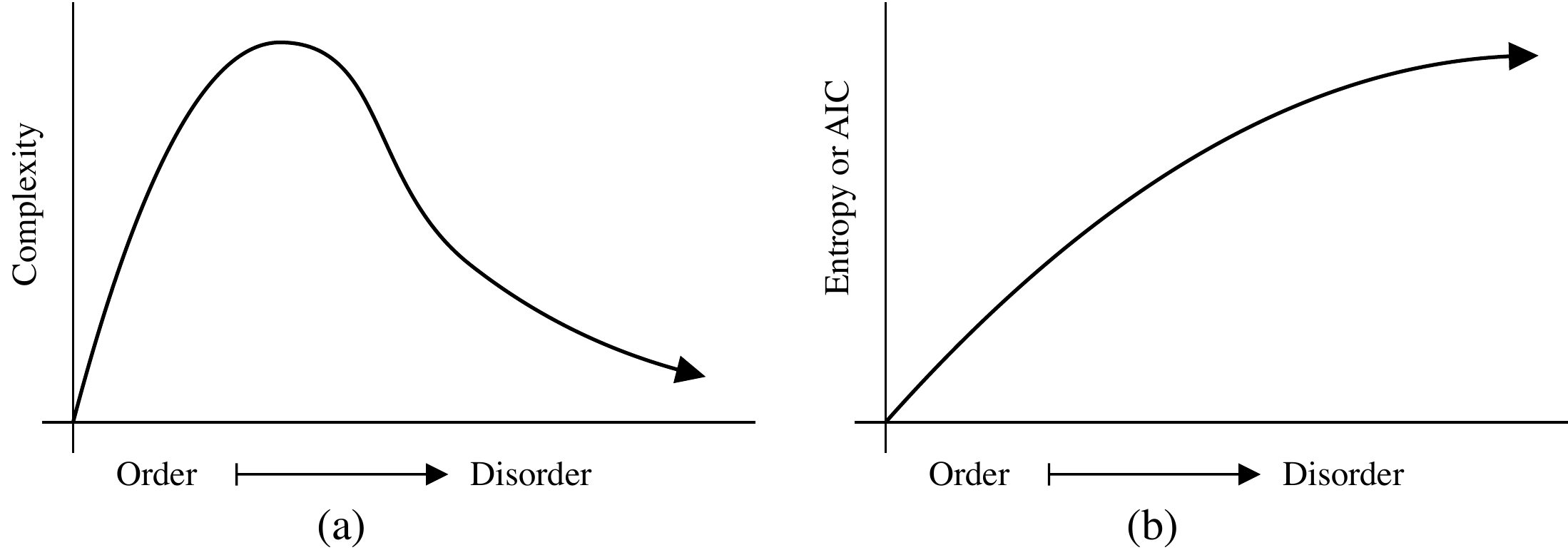}
  \caption{Plots of (a) complexity versus disorder and (b) entropy or
  algorithmic information content (AIC) versus disorder. The complexity peaks
  somewhere between order and disorder while both the entropy and AIC grow
  monotonically with the disorder of the system.}
  \label{fig:measure_vs_disorder}
\end{figure}

An alternative approach is to use a single quantitative measure of complexity,
of which many have been proposed~\citep{lloyd2001measures,
mitchell2009complexity}. Figure~\ref{fig:measure_vs_disorder} demonstrates the
typical challenge of measuring complexity. In
figure~\ref{fig:measure_vs_disorder}a, we see that the most complex systems lie
between order and disorder. Common measures of information content such as
Shannon entropy~\citep{shannon1948mathematical} or algorithmic information
content (AIC; also known as Kolmogorov complexity)~\citep{kolmogorov1965three,
solomonoff1964formal} grow monotonically with the disorder of the system (fig.
\ref{fig:measure_vs_disorder}b).

Several complexity measures have been proposed which in theory peak in the zone
of complexity between order and disorder. These include measures such as
effective complexity~\citep{gell1996information}, minimum message
length~\citep{wallace2005statistical}, statistical
complexity~\citep{crutchfield1989inferring} and the closely related effective
measure complexity~\citep{grassberger1986toward}. One problem with these
measures is that they are not easy to compute. They typically require computing
the AIC or the entropy (or related information-theoretic measures) over a large
state space. Also, they often require assumptions on the part of the modeler,
which degrades their objectivity. Few, if any, practical examples exist which
use these measures to compute the complexity of a dynamical system.

Statistical complexity is perhaps the most objective as it does not require any
data-specific assumptions to put in practice. \citet{shalizi2004quantifying}
used statistical complexity to separate circular cellular automata exhibiting
sophisticated structure from those which are more ordered and more disordered.

Alternatively, compression-based or prediction-based heuristics can be used to
measure complexity~\citep{cisneros2019evolving, zenil2010}. These heuristics
were able to correctly order by complexity Wolfram's four classes of elementary
cellular automata~\citep{wolfram2002new}.

Completely reducing the identification of intelligent behavior or even
complexity to a single measure is impractical. Other tools to assist in the
scalable identification of these sorts of behaviors will be useful.  These
include producing summary visualizations~\citep{cisneros2020visualizing}, and
filtering large state spaces for interesting
structures~\citep{hanson1997computational}.

%% file: redistribution.tex
\section{A Redistribution}

I began by making the case that researchers should investigate more resources
in the phylogeny of machine intelligence. Human-like
intelligence is hard to design. Automating the process through phylogenesis may
be both faster and more objective. Observing nature, I noted that the bulk of
the computation which went into producing humans is found in phylogenesis over
ontogenesis. While the type of computation may differ in value-per-operation,
the orders of magnitude more invested in phylogenesis is suggestive. 

To investigate more resources in phylogenesis, we must first consider the
options. These make up the phylogenetic stack (which is really a spectrum)
beginning with emergent evolution and culminating with machine learning.

I assessed each possible starting point from several perspectives. These
included the trade-offs between complexity and computation, the trade-offs of
assumptions and computation, and the likelihood of intelligent behavior
emerging from a given starting point. Throughout, I used nature as a crutch in
my analysis. This may be a major flaw, but is unavoidable if we care to make
any inductions versus purely deductive speculation. After all, nature is the
only system that we know of which has evolved human-like intelligence.

Given the above analysis, determining a starting point in the phylogenetic
stack for evolving intelligence \emph{in silico} remains non-trivial. However,
we need not determine precisely where to start since we need not limit
ourselves to a single choice. Rather, a more interesting question is how to
distribute the resources for research across the phylogenetic stack. My 
primary conclusion is that we should devote more resources to earlier in the
stack.

Within phylogenesis, meta-evolutionary algorithms are well supported as a
starting point. They do not suffer the limitations of EAs which
have been observed in practice. They also retain many of the benefits of
purely emergent evolution while avoiding many of the contingencies.
Regardless, by studying simulations of evolution, we will undoubtedly learn
more about how to develop machine intelligence -- and maybe even a clue or two
about the nature of ourselves.